\documentclass{article}
\usepackage[nonatbib, final]{pietro}

\usepackage[utf8]{inputenc} 
\usepackage[T1]{fontenc}    
\usepackage{hyperref}       
\usepackage{url}            
\usepackage{booktabs}       
\usepackage{amsfonts}       
\usepackage{nicefrac}       
\usepackage{microtype}      
\usepackage{amsmath,amsfonts,amssymb}
\usepackage{graphicx}
     
\title{Correlation Alignment by Riemannian Metric for Domain Adaptation}

\author{
  Pietro Morerio and Vittorio Murino\\
  Pattern Analysis and Computer Vision (PAVIS)\\
  Istituto Italiano di Tecnologia\\
  via Morego, 30 - 16163 Genova, Italy \\
  \texttt{name.surnameo@iit.it} \\
}

\begin{document}

\maketitle

\begin{abstract}
Domain adaptation techniques address the problem of reducing the sensitivity of machine learning methods to the so-called domain shift, namely the difference between source  (training) and target (test) data distributions. In particular, unsupervised domain adaptation assumes no labels are available in the target domain. To this end, aligning second order statistics (covariances) of target and source domains have proven to be an effective approach ti fill the gap between the domains. However, covariance matrices do not form a subspace of the Euclidean space, but live in a Riemannian manifold with non-positive curvature, making the usual Euclidean metric suboptimal to measure distances. In this paper, we extend the idea of training a neural network with a constraint on the covariances of the hidden layer features, by rigorously accounting for the curved structure of the manifold of symmetric positive definite matrices. The resulting loss function exploits a theoretically sound geodesic distance on such manifold. Results show indeed the suboptimal nature of the Euclidean distance. This makes us able to perform better than previous approaches on the standard Office dataset, a benchmark for domain adaptation techniques.
\end{abstract}

\section{Introduction and motivation}
\label{sec:intro}
Deep Neural networks are able to learn powerful hierarchical representations from large training sets and show very good generalization capabilities. Furthermore, learned features are so general that can often be successfully transferred across domains and tasks, especially when one has the possibility to fine tune the network exploiting labelled examples from the new domain.
Yet, deep architectures are not completely immune from the so called \textit{domain shift} problem \cite{un_bias}, i.e. they suffer from performance degradation under changes in the input data distribution \cite{Donahue_ICML2014}. This is what typically happens when training supervised machine learning algorithms to be deployed in ``the wild'', where test (\textit{target}) distributions can be extremely different from the training one (\textit{source}), and labeled test/validation samples are usually not available. This issue is known under the name of \textit{domain adaptation} (DA) and addressing it, especially in the unsupervised case, is in fact critical for successfully applying machine learning to real-world applications. 

Actually, one would like to avoid collecting and labeling data to train a new classifier for any new possible scenario, and in some real cases this is not actually possible. Instead, it would be desirable to find methods that cope with the degradation in classification performance by effectively transferring the knowledge acquired on the labeled source domain to some unlabeled target domain \cite{Fernando}.

To cope with this issue, a possibility consists in minimizing the distance between source and target data marginal distributions so as to estimate a transformation of the feature representations which may lead to better classification. In other words, this is also equivalent to \textit{confuse} the domains so that a classifier cannot distinguish between source and target domains.

Many DA works tackle this problem according to this intuition and propose methods aimed at \textit{aligning} information extracted from the domains' data, be either lower dimensional manifolds, subspaces or distributions \cite{Fernando, CORAL}
This alignment can be performed in an unsupervised fashion, although it can strongly benefit from some labeled samples \cite{Tzeng_ICCV2015}.

In this context, a recent paradigm in unsupervised DA proposes to align second-order statistics of source and target distributions. This method is named CORAL (CORrelation ALignment) \cite{CORAL} and proposed a ``frustratingly easy'' unsupervised domain adaptation method by finding the linear transformation which minimizes the Frobenius norm of the difference between the covariance matrices of source and target data features. 



Although straighforward, the method looks effective, however, it shows two main drawbacks. First, it relies on a linear transformation, which could be insufficient to capture the most appropriate feature transformations. 
In fact, the way features are (cor)related across the domains is not known, and assuming a linear relations is big bet, especially if deep feature representations (e.g., from last fully connected layers, \textit{fc7} and/or \textit{fc8}) are considered.
Second, it is not end-to-end since it needs to extract features from both source and target datasets, calculate covariances, align the features, and then train a classifier. 

Indeed, the latter drawback of CORAL was recently addressed by Deep CORAL \cite{DeepCORAL}, which incorporates the alignment of second-order statistics into a deep architecture, by proposing a loss term which minimizes the batch-wise difference between source and target data correlations. In essence, Deep CORAL aims at optimizing the weights of a deep architecture jointly considering the optimization problem of the covariance difference and the standard classification problem. This is done by designing a loss composed by the standard (cross-entropy) classification loss and another, properly weighted, loss penalizing the covariances' difference.



This paper operates just in this context by addressing a fundamental issue inherent of the above approaches.
Although CORAL and Dep CORAL showed good results and proved effective, they overlooked fundamental properties of covariances, which turn out to be Symmetric Positive Definite (SPD) matrices. A key property of such matrices is that, given $n \in \mathbb{N}$, the set of $n \times n$ SPD matrices is not a subspace of the Euclidean space, but instead has the structure of a Riemannian manifold with non-positive curvature, usually denoted as $\mathrm{Sym}^{++}(n)$. 
As a consequence, methods for manipulating elements in $\mathrm{Sym}^{++}(n)$ which simply rely on the Euclidean metric 
are usually \textit{suboptimal} \cite{NIPS2014_5457}. This is quite intuitive, since the Frobenius norm of a matrix difference, $\| A - B \|_F $, is defined only in terms of the element-wise difference $A-B$, without reflecting any structure in $A$ and $B$.


Many methods have been proposed in the literature which exploit the non-Euclidean nature of $\mathrm{Sym}^{++}(n)$ \cite{Dryden}.

A common approach exploits the affine-invariant metric, a classical Riemannian metric on $\mathrm{Sym}^{++}(n)$ \cite{Conversano2012605, Pennec2006}, which is typically computationally intensive, particularly when large-scale applications are faced.
Another approach exploits Bregman divergences on $\mathrm{Sym}^{++}(n)$ \cite{Kulis06learninglow-rank, journals/pami/CherianSBP13}. These are not Riemannian metrics but are quite fast to compute and proved to work properly on retrieval tasks, even considering very simple classification methods such as nearest-neighbor methods.
The computational complexity is undoubtely one of the main drawback affecting the manipulation of such objects which limits the usage of such tools, despite their elegant and rigorous mathematical soundness. This motivated the development of the \textit{Log-Euclidean metric} framework \cite{Arsigny:Siam:07, conf/cvpr/WangGDD12}, which is faster than the affine-invariant metric and, moreover, is a Riemannian metric on $\mathrm{Sym}^{++}(n)$ (unlike the Bregman divergences), and thus can better suits its manifold structural form. 


The latter distance was recently exploited in computer vision and machine learning tasks \cite{Minh:CVPR16,cavazza2016kercov} since it is fast to compute, and in particular, has the interesting property of being  differentiable. In our case, this property permits to compute gradients with respect to each entry of the source and target covariance matrices allowing end-to-end optimization via gradient descent techniques.

\paragraph{Our contribution.}
In this work, we show that, leveraging the Riemannian structure of $\mathrm{Sym}^{++}(n)$, domain adaptation can be performed in a more effective and principled way. In fact, second order statistics must be properly aligned within their natural embedding manifold, instead of naively projected in the Euclidean space.
Since the Euclidean distance is proved to be suboptimal on the curved manifold $\mathrm{Sym}^{++}(n)$, we introduce a loss function based on the Log-Euclidean metric, introducing \textit{de facto} a novel, more rigorous version of the Deep CORAL framework. 
This allows to effectively and correctly align second order statistics of the source and target data domains, whose effect results even more evident whenever source and target datasets are characterized by very different marginal distributions. Experiments performed on a standard domain adaptation benchmark, the Office dataset, show superior performance, empirically confirming the correctness of our approach. As a side finding, we notice a pathological behavior of the Euclidean distance, which can be interpreted as a symptom of its sub-optimality with respect to geodesic distances in $\mathrm{Sym}^{++}(n)$.


\paragraph{Organization of the paper.} 
The next section surveys and discusses existing methods in unsupervised domain adaptation. Section \ref{sec:cov} contains some necessary background material on covariance matrices and manifold distances, which lay the ground for section \ref{sec:log-loss}, which addresses the core methodological contribution of the paper. Empirical results are provided and discussed in section \ref{sec:res}, while section \ref{sec:concl} draws some conclusions and proposes possible extensions of the work.

\section{Related Work}\label{sec:relwork}
Domain adaptation techniques address the fundamental problem of \textit{domain shift} \cite{un_bias} between a \textit{source} dataset, used for training and a \textit{target} dataset, used for testing. This issue often arises in machine learning, especially when algorithms are to be deployed on new, possibly unlabeled domains. Domain adaptation strategies can be divided in two classes: \textit{supervised} and \textit{unsupervised} adaptation. The first approach is based on the assumption that a supervised algorithm (usually a classifier) can benefit not only from a fully labeled source dataset, but also from (at least) some labeled data points from the target dataset. Our method belongs instead to the second class of approaches, which assume that no labels are available for the target data. In the following we detail recent methods proposed in literature to cope with this problem.

A first class of methods aims to learn transformations which somehow align feature representations in the source and target sets. For instance, in \cite{Glorot11domainadaptation} auto-encoders are exploited to learn common features. In \cite{bishift}, a bi-shifting auto-encoder (BSA) is instead intended to shift source domain samples into target ones. \cite{Dict1,Dict2} approach the problem relying on dictionary learning, to find representations where source and target datasets are aligned. Geodesic methods \cite{Geod1,Geod2} aim  at projecting source and target datasets on a manifold, and connecting the two subspaces with a path. Then, both datasets can be projected along this path. Eventually, \cite{CORAL,Fernando} propose a transformation to minimize the distance between the covariances of source and target datasets.

As a second paradigm, recently the application of deep neural networks for domain adaptation has been investigated, to reduce the domain shift through end-to-end training. In \cite{DLID}, an interpolating path between the two domains is defined, and features are extracted in unsupervised manner from the interpolating domains. Such features are further combined to feed a classifier. In \cite{DDC} an adaptation layer and an additional domain adaptation loss are added to a standard CNN architecture. In \cite{DANN} the mismatch between source and target domains is tackled by reducing the Maximum Mean Discrepancy between the two. In \cite{Ganin} the two domains' representations are aligned by adding an adversarial loss aimed at making a neural network confusing the two domains, thus learning features which are representative for both. Backpropagation algorithm and a gradient reversal layer are used to train the so-defined model. In \cite{RKHS} hidden representations are explicitly matched in Reproducing Kernel Hilbert Spaces. In \cite{Beyond} a two-stream neural network is proposed, where one is dedicated to modeling the source samples and the other to modeling the target ones, without weight sharing. In \cite{Transductive} the representation, the transformation between the two domains and the target label inference are optimized in a end-to-end fashion, exploiting transductive learning. In \cite{DeepCORAL}, the second order statistics of source and target deep representations are aligned, through the addition of a CORAL loss term to the overall loss. Our work is a direct extension of the latter approach as discussed in details in section \ref{sec:log-loss}.

A different deep learning based approach exploited for unsupervised domain adaptation relies on Generative Adversarial Networks. In \cite{Coupled_GANS}, the joint distribution between the two domains is learned, while \cite{cross_domain_image} directly finds a way to transform samples from the source distribution into samples from the target distribution (and viceversa). These approaches seem very promising, though not fully tested in all the benchmark datasets.

Last worth to mention is Batch Normalization \cite{batch_norm}, recently revisited under a specific domain adaptation perspective \cite{li2016revisiting}. The idea is quite similar in spirit to the idea of aligning second order statistics. However Batch Normalization tries to compensate for the source's internal covariance shift by normalizing each mini-batch to be zero-mean and unit variance (whitening), while not taking into account target's statistics: layer's activation are decorrelated for the source data, but not for target points.

\section{Background: covariance matrices and manifold distances}
\label{sec:cov}

Covariance descriptors have been widely explored in computer vision for a plethora of task, ranging from tracking \cite{Porikli2006}, object detection \cite{TPM:ECCV06}, image classification \cite{Harandi:CVPR14,Minh:CVPR16} and action recognition \cite{cavazza2016kercov,Camps:CVPR16}.

Covariance operators, once properly regularized, are symmetric and positive definite operators which does not form a vectorial space, but a Riemannian manifold instead. For this reason, Euclidean distance is suboptimal to measure distances since it does not consider the inner curvature of the ambient space. As an alternative, \emph{geodesic distances} are actually exploited since able to compute the shortest path on the manifold. Surely, the affine-invariant metric
\begin{equation}\label{eq:dA}
d_{a}(A,B) = \|\log(A B^{-1})\|_F,
\end{equation}
is the most common one. Similarly, symmetrized Bregman divergences \cite{Harandi:CVPR14} are frequently used to measure distances on the manifold: among them, we can mention the Jeffrey divergence \cite{Moakher:VPTF:06} 
\begin{equation}\label{eq:dJ}
d_{J}(A,B) = \dfrac{1}{2} {\rm tr}(A^{-1}B) + \dfrac{1}{2} {\rm tr}(B^{-1}A) - n,
\end{equation}
for any $A,B \in {\rm Sym}^{++}(n)$. Although $d_J$ is not a proper distance (it does not fulfill the triangular inequality), it is commonly used \cite{Camps:CVPR16} as well as the Stein divergence \cite{Cherian:PAMI:13}
\begin{equation}\label{eq:dS}
d_{S}(A,B) = \sqrt{ \log \left| \dfrac{A + B}{2} \right| - \dfrac{1}{2} \log | AB | }.
\end{equation} 
Since $d_a$, $d_{J}$ and $d_{S}$ are explicitly defined in terms of matrix multiplications and because $d_a$ and $d_{J}$ require to compute an inverse, they does not scale well with respect to $n$, the dimension of $A$ and $B$. In order to mitigate this drawback, the Log-Euclidean metric \eqref{logE} is very convenient since just defined as the Frobenius distance of the logarithms. 


\section{Log-Euclidean distance for deep correlation alignment}
\label{sec:log-loss}

CORAL (CORrelation ALignment) \cite{CORAL} is an unsupervised DA method which consists in aligning second-order statistics of source and target data distributions (typically, after normalization and zero-mean transformations). 
It finds the linear transformation $A$ which minimizes the Frobenius norm $\| \cdot \|_F$ of the difference between the covariance matrices of source and target data, $C_S$ and $C_T$ respectively, by solving:
\begin{equation}\label{eq:CORAL}
\min_A \| C_{\hat{S}} - C_T \|^2_F = \min_A \| A^T C_S A - C_T \|^2_F.
\end{equation}
The transformation $A^*$ which, acting on the source data, minimizes (\ref{eq:CORAL}), has essentially the form of a whitening operation, followed by a re-coloring of the whitened features, performed through the covariance operator of the target domain. It can be applied in any kind of DA problem, regardless of the chosen features (and also deep features can be used) and classification method used afterwards. 

The minimization problem (\ref{eq:CORAL}) is analytically solved to provide the optimal solution \cite{CORAL}:
\begin{equation}\label{eq:A-star}
A^* = (U_S {\Sigma_S^+}^{\frac{1}{2}} U_S^{\mathsf{T}}) (U_{T[1:r]} \Sigma_{T[1:r]}^{\frac{1}{2}} U_{T[1:r]}^{\mathsf{T}} ).
\end{equation}
Here $r$ is the minimum rank of the source and target covariance, $r = \min(r_{C_s},r_{C_T})$, $U_S \Sigma_S U_S^{\mathsf{T}}$ is the diagonalization of $C_S$, $\Sigma_S^+$ is the Moore-Penrose pseudoinverse of $\Sigma$ and $U_{T[1:r]} \Sigma_{T[1:r]} U_{T[1:r]}^{\mathsf{T}}$ are the largest $r$ singular values and corresponding vectors resulting from the diagonalization of $C_T$ . $A^*$ is cleary made by a first part, which whitens the source data and a second one which re-colors it with the target statistics. However, in practice, for the sake of efficiency and stability, CORAL employs the standard whitening and recoloring, where a small regularization term $\gamma \mathbb{I}_d$ is added to the covariance matrices in order to make them explicitly full rank and positive definite.

Deep CORAL \cite{DeepCORAL} incorporates the alignment of second-order statistics into a deep architecture, by proposing a loss term which minimizes the batch-wise difference between source and target correlations. Deep CORAL indeed aims at optimizing the weights of a deep architecture by jointly solving the problem (\ref{eq:CORAL}) and the standard classification problem, designing by a compound loss so composed:
\begin{equation}
L = L_{CLASS} +  \lambda L_{CORAL},
\end{equation}
where
\begin{equation}\label{eq:D-CORAL}
L_{CORAL} = \frac{1}{4d^2} \| C_S - C_T \|^2_F.
\end{equation}
and $L_{CLASS}$ is the standard cross-entropy loss. $L_{CORAL}$ is calculated (for each batch) over the $d$-dimensional \textit{fc8} features of AlexNet \cite{AlexNet}, but according to \cite{DeepCORAL} could possibly include contributions from hidden representations at each level of the network. This formulation permits to compute gradients with respect to each entry of the source and target covariance matrices allowing end-to-end optimization via gradient descent techniques.

However, covariance (SPD) matrices live in a Riemannian space $\mathrm{Sym}^{++}(n)$, and metrics defined therein should take into account its non-Euclidean structure, so the (Euclidean) distance present in (\ref{eq:D-CORAL}) is only suboptimal in such a space. 
The \textit{Log-Euclidean metric} is instead a Riemannian metric and better captures the manifold structure. It is characterized by some interesting properties (which will be illustrated in the following) and is defined as:
\begin{equation}\label{logE}
d_{logE} (X,Y) = \| \log(X) - \log(Y) \|_F,
\end{equation}
where $\log(A)$ is defined as $\log(A)=U \mathrm{diag}(\log(\lambda_1), ... ,\log(\lambda_n))U^{\mathsf{T}}$ through the spectral decomposition $A=U \mathrm{diag}(\lambda_1, ... ,\lambda_n)U^{\mathsf{T}}$.

In the context of CORAL, it is straightforward to note that the solution of problem (\ref{eq:CORAL}), $A^*$, incidentally minimizes the analogous problem, that is:
\begin{equation}\label{eq:log-CORAL}
\min_A \| \log(C_{\hat{S}}) - \log(C_T) \|^2_F = \min_A \| \log(A^T C_S A) - \log(C_T) \|^2_F.
\end{equation}

Actually, supposing to work with the regularized full-rank matrices (which is indeed what people do in practice), the minimization problem (\ref{eq:CORAL})  finds the transformation $A^*$ which makes the Frobenius distance \textit{null}, i.e., the transformation which realizes the equality $C_{\hat{S}} = C_T$. This of course also yields  $\log(C_{\hat{S}}) = \log(C_T)$, which minimizes eq. (\ref{eq:log-CORAL}) as a consequence of the fact both metrics are well defined (in either spaces, they comply with distance properties), and satisfy  $d(X,Y)=0 \Leftrightarrow X=Y$. Namely, what matters in CORAL is only the analytical function which directly transforms the data to make the two covariances \textit{the same}. 

But the Deep CORAL \cite{DeepCORAL} methods works quite differently. In this case, domain adaptation is end-to-end, and the problem is addressed by jointly optimizing the loss for the supervised problem and the Euclidean loss in eq. (\ref{eq:D-CORAL}). The transformation $A^*$ is now implicitly learned step-by-step by the network via gradient descent. In other words, the final deep features are both discriminative enough to train a strong classifier and invariant (to some extent) to the difference between source and target domains. However, the minimization is now a smooth process, which follows a precise path in the parameter (weight) space. Such a path naturally induces a trajectory in $\mathrm{Sym}^{++}(n)$ which connects the covariance of the source with the one of the target. It is thus natural to constrain such a path to be a \textit{geodesic} trajectory, enforcing a minimum distance which takes into account the curvature of the manifold $\mathrm{Sym}^{++}(n)$. 

As a side consideration, let us note that a \textit{perfect} alignment of the source and target distributions up to second order statistics is indeed a very strong assumption done in CORAL \cite{CORAL}. A more reasonable and milder constraint is to have a balance between good features in the source domain and a sound statistical adaptation to the target distribution.

\paragraph{The LOG-D-CORAL loss.}
Based on the above considerations, and on the remarks of section \ref{sec:cov}, we propose to address the unsupervised domain adaptation problem by adding to a deep network a loss term based on a geodesic distance on $\mathrm{Sym}^{++}(n)$, namely the log-Euclidean one, which, as already mentioned, offers theoretical and practical advantages over other metrics on $\mathrm{Sym}^{++}(n)$.
\begin{eqnarray}\label{eq:logLoss1}
L_{log} & =  & \frac{1}{4d^2} \| \log(C_S) - \log(C_T) \|^2_F \\
&= &\frac{1}{4d^2} \| U \mathrm{diag}(\log(\lambda_1), ... ,\log(\lambda_d))U^{\mathsf{T}} - V \mathrm{diag}(\log(\mu_1), ... ,\log(\mu_d))V^{\mathsf{T}} \|^2_F  ,\label{eq:logLoss2}
\end{eqnarray}
where $d$ is the dimension of the hidden features whose covariances are intended to be aligned, $U$ and $V$ are the matrices which diagonalize $C_S$ and $C_T$, respectively, and $\lambda_i$ and $\mu_i, \; i=1,...,d$ are the corresponding eigenvalues. The normalization term $1 / d^2$ accounts for the sum of the $d^2$ terms in the Frobenius distance, which makes the loss independent from the size of the features.

Jointly training with a standard classification loss and the proposed loss in (\ref{eq:logLoss1}) allows to learn features which do not overfit the source data since they reflect the statistical structure of the target set. Hence, the total loss reads
\begin{equation}
L = L_{CLASS} +  \alpha L_{log}.
\end{equation}

The hyperparameter $\alpha$ is a critical coefficient. A high value of $\alpha$ is likely to force the network towards learning oversimplified low-rank feature representations, which may have perfectly aligned covariances but would be useless for classification purposes. On the other hand, a small $\alpha$ may not be enough to fill the domain shift.

\paragraph{Differentiability.}
The loss (\ref{eq:logLoss2}) needs to be differentiable in order for the minimization problem to be solved via back-propagation, and its gradients should be calculated with respect to the input features. 
Given a zero mean data matrix $D \in \mathbb{R}^{L\times d} $, composed by $L$ samples of $d$ dimensional vectors, its covariance is simply proportional to the quadratic form $D^{\mathrm{T}}D$, whose gradients can be straightforwardly computed. 

The scenario is indeed more complicated than expected since the logarithm of an SPD matrix is defined through its eigendecomposition in eqs. (\ref{eq:logLoss1}) and (\ref{eq:logLoss2}). Fortunately, eigenvalues and eigenvectors are differentiable functions for SPD matrices \cite{Kriegl2003}. Lastly, the point-wise $\log$ is applied in (\ref{eq:logLoss2}) on the matrix listing  \textit{strictly positive} eigenvalues\footnote{Remember that covariances are in practice regularized by adding a small perturbation $\gamma \mathbb{I}$.} on the diagonal, and it is thus differentiable everywhere as a function of $\lambda_i$ and $\mu_i$.

In practice, modern tools for deep learning consist in software libraries for numerical computation whose core abstraction is represented by \textit{computational graphs}. Single mathematical operations (e.g., matrix multiplication, summation etc.) are deployed on nodes of a graph and data flows through edges. Reverse-mode differentiation \cite{griewank2012invented} takes advantage of the gradients of single operations, allowing training by backpropagation through the graph \cite{Colah}. The loss (\ref{eq:logLoss2}) can be easily written in few lines of code by exploiting mathematical operations already implemented, together with their gradients, in TensorFlow\texttrademark \cite{TF}.

\section{Results}
\label{sec:res}

\begin{figure}[h]
	\centering
	\includegraphics[width=0.18\textwidth]{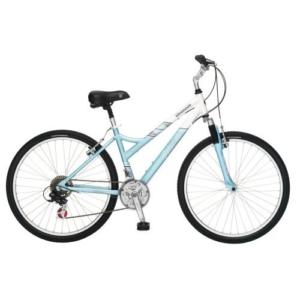}
	\includegraphics[width=0.18\textwidth]{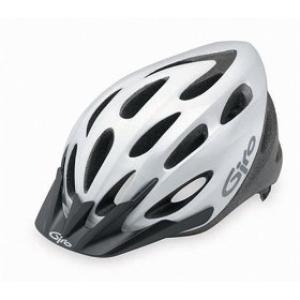}
	\includegraphics[width=0.18\textwidth]{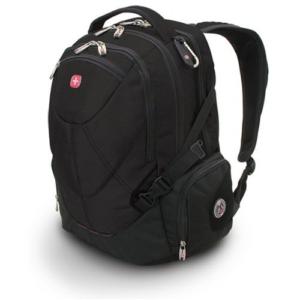}
	\includegraphics[width=0.18\textwidth]{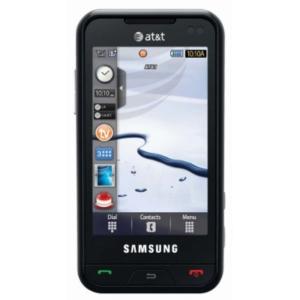}
	\includegraphics[width=0.18\textwidth]{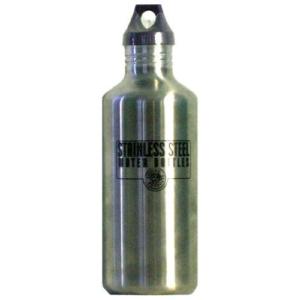} \\
	\includegraphics[width=0.18\textwidth]{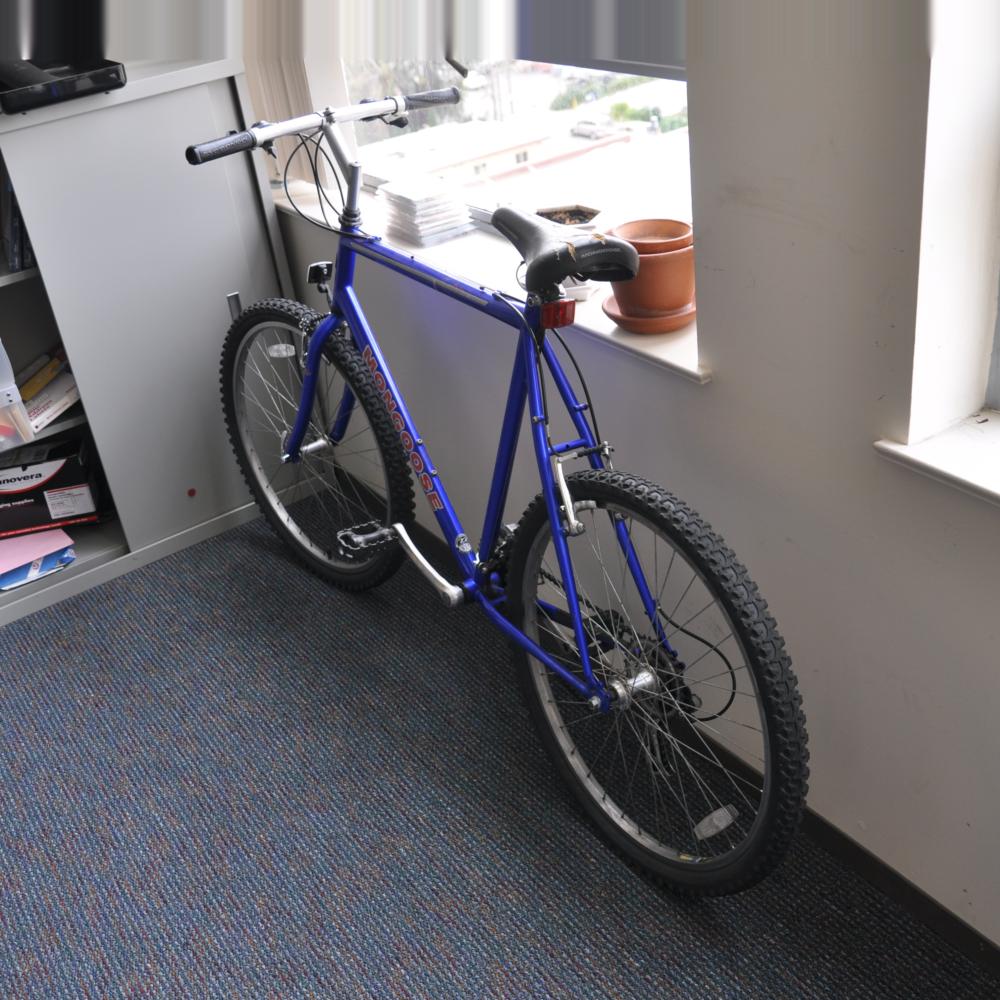}
	\includegraphics[width=0.18\textwidth]{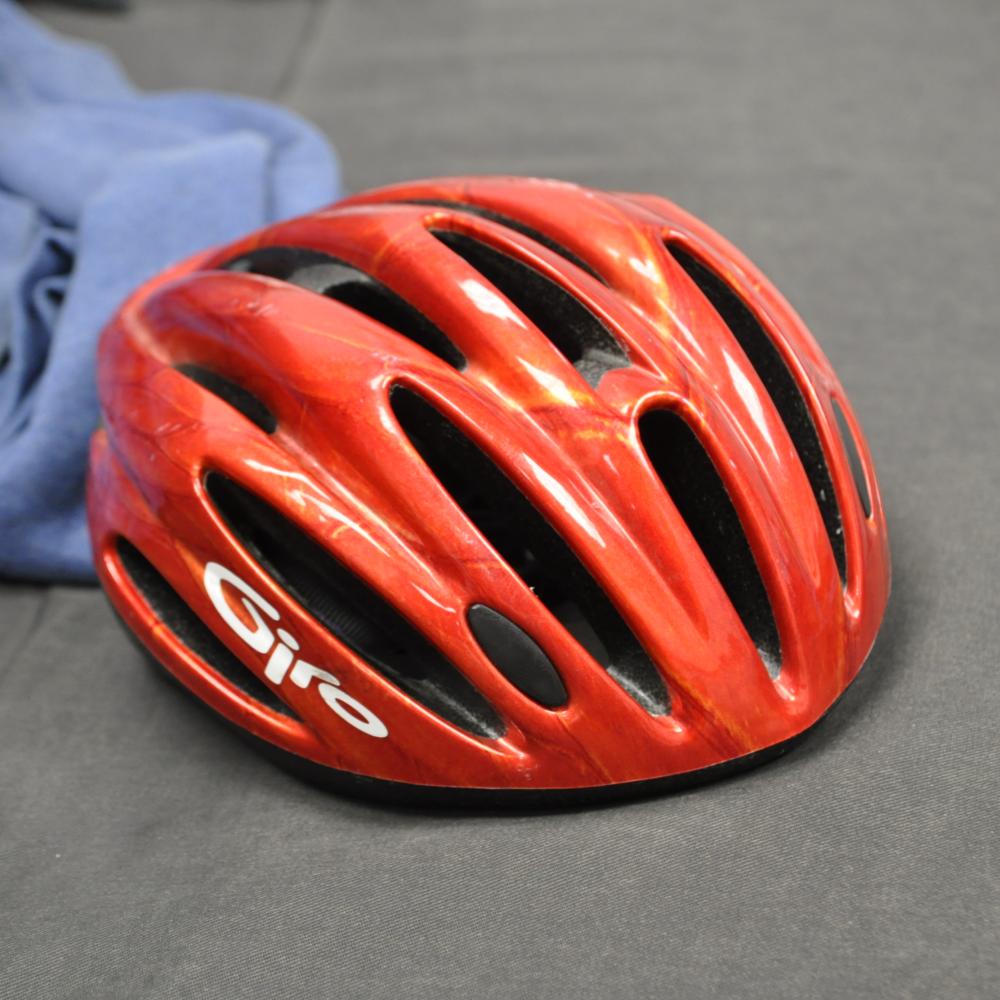}
	\includegraphics[width=0.18\textwidth]{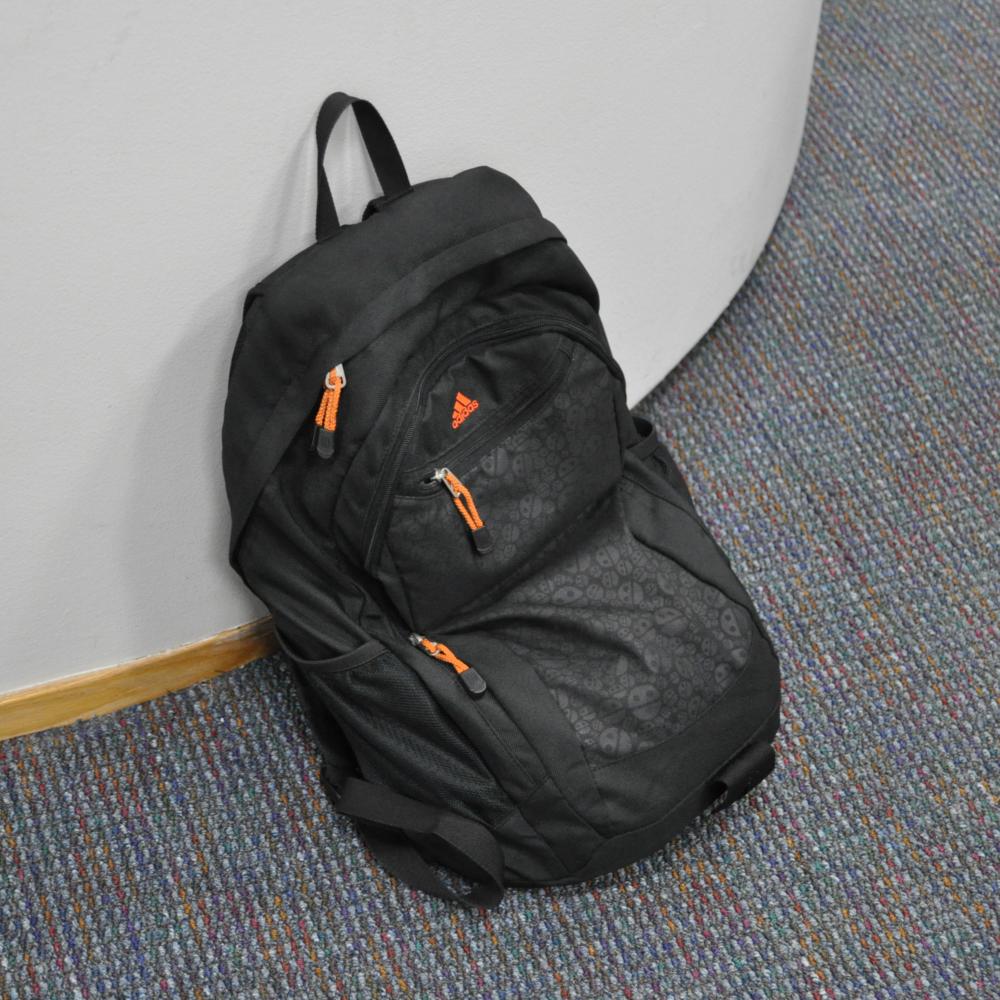}
	\includegraphics[width=0.18\textwidth]{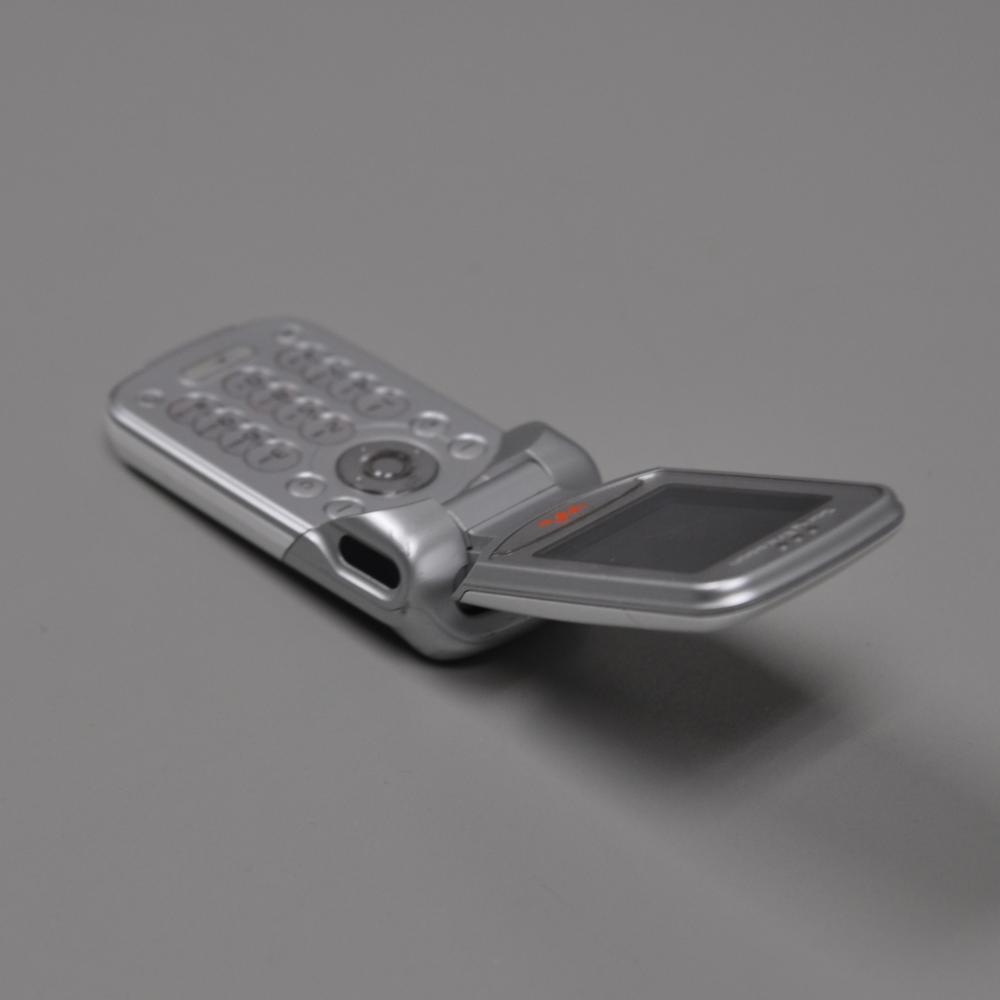}
	\includegraphics[width=0.18\textwidth]{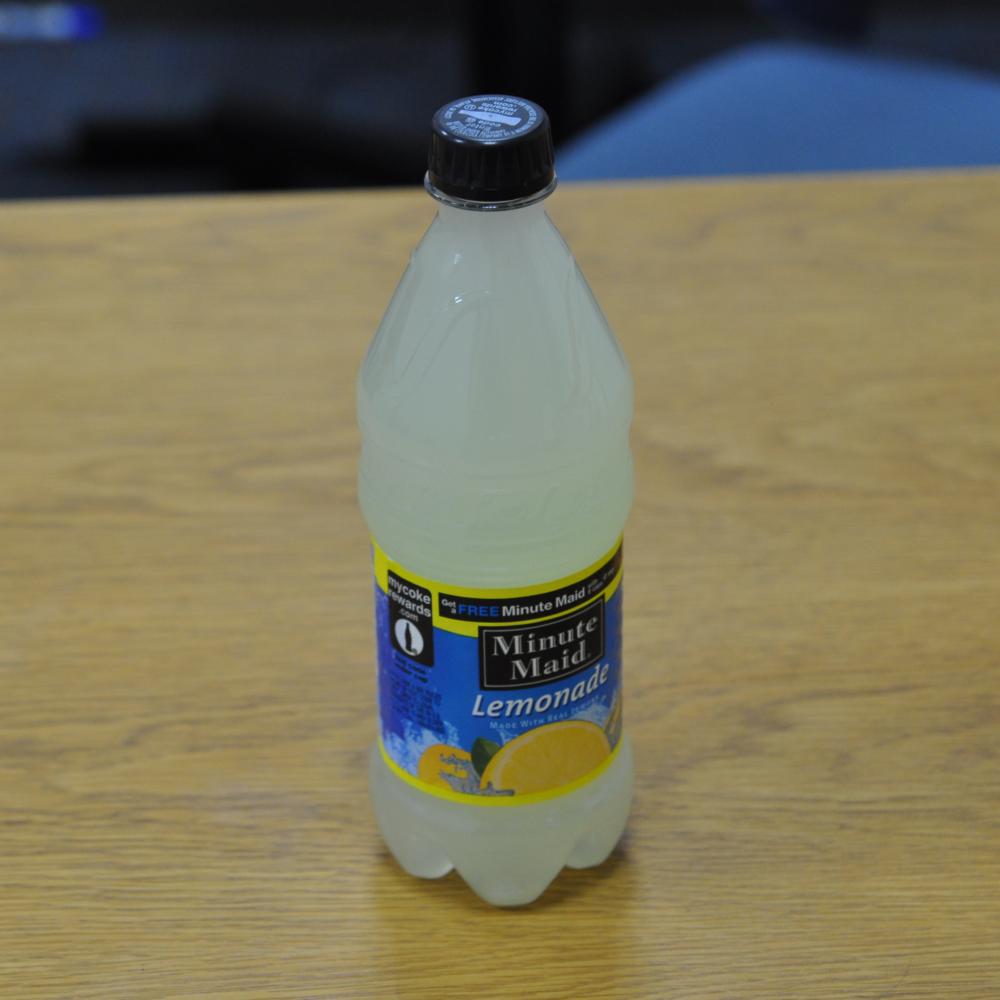} \\
	\includegraphics[width=0.18\textwidth]{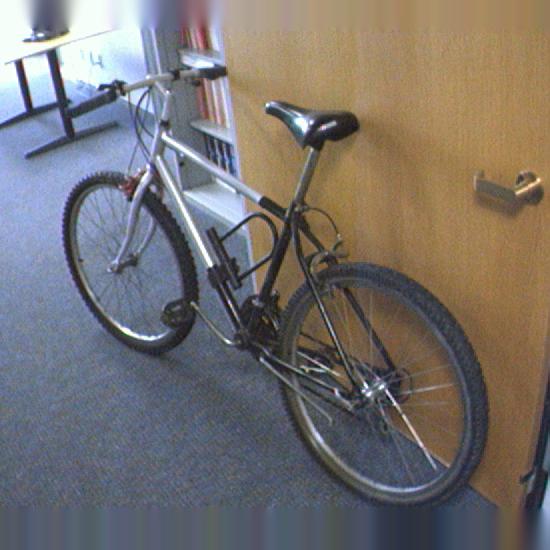}
	\includegraphics[width=0.18\textwidth]{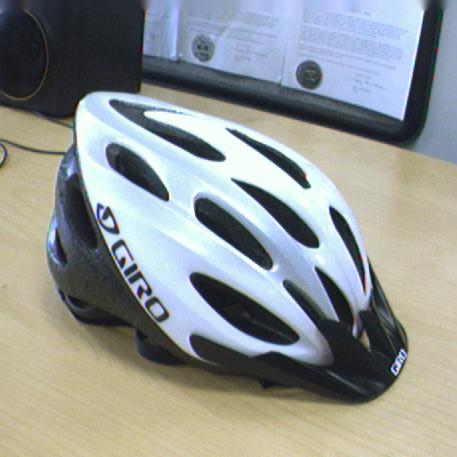}
	\includegraphics[width=0.18\textwidth]{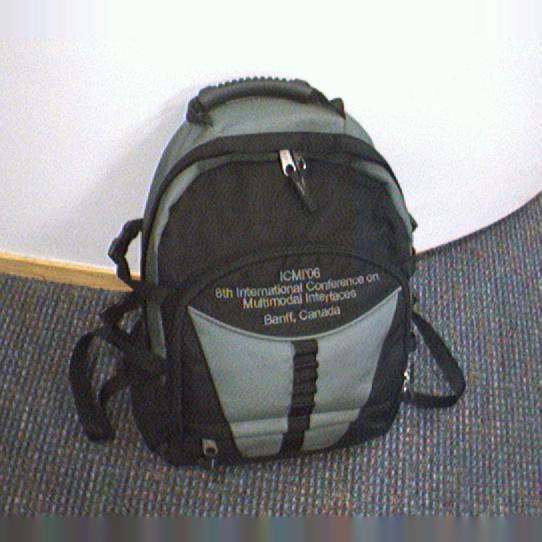}
	\includegraphics[width=0.18\textwidth]{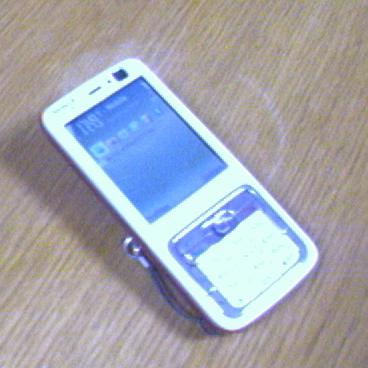}
	\includegraphics[width=0.18\textwidth]{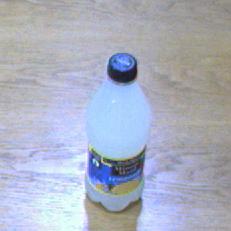}
	\caption{Samples from 5 classes of the Office \cite{Office} dataset. Each row correspond to one of the three different domains, namely Amazon, DSLR, Webcam (top to bottom). Similarity between DSLR and AMAZON reflects in very high recognition accuracies in the evaluation setups W $\rightarrow$ D and D $\rightarrow$ W, as reported in table \ref{tab:results}. In fact, the effect of domain adaptation techniques are in general more evident on the the other domain shifts.}
	\label{fig:office}
\end{figure}

We reproduce the experimental setup proposed in Deep Coral \cite{DeepCORAL}, by validating our approach on a the standard domain adaptation benchmark - the Office dataset \cite{Office}. The dataset consists in images belonging to 31 different object categories, gathered from 3 different domains, namely \textit{Amazon}, \textit{DSLR} and \textit{Webcam} (Figure \ref{fig:office}). The most standard evaluation protocol for unsupervised domain adaptation (see e.g.\cite{Geod2,DDC,Ganin}) is simple: given the three domains, there are 6 possible domain shifts. Training is performed on the the fully labeled data from a given source domain, while testing is done over the two remaining (targets). Only \textit{unlabeled} data from the current target domain is available at training time, which allows to work out statistics to fill the domain gap.

\paragraph{Implementation details.}
We fine tune AlexNet \cite{AlexNet} pre-trained on Imagenet \cite{Imagenet}, by setting the dimension of its last hidden layer (\textit{fc8}) to 31, i.e. the number of classes in the Office dataset. The new layer is initialized with Gaussian noise $\mathcal{N}(0, 5\times 10^{3})$, batch size is 128 and base learning rate $10^{-3}$, with scheduled exponential decreasing. Batches are made of both target and source examples, where the former contribute to $L_{log}$ or $L_{CORAL}$ losses only, while the latter (which are labeled) also contribute to the cross entropy loss $L_{CLASS}$. The network is trained separately on the three domains and tested on the remaining ones, to serve as a baseline (first row of Table \ref{tab:results}). Unfortunately we were not able to reproduce the performance of AlexNet reported by \cite{DeepCORAL} by some percentage points, although we accurately followed all of its prescriptions. This does not really matter since we are only interested in the \textit{relative gain} in performance introduced by our loss function with respect to the suboptimal Euclidean one. Loss weights $\alpha$ and $\lambda$ used for each domain shift are listed in Table \ref{tab:parameters}, while the covariance regularizer $\gamma$ was set to $10^{-5}$ once for all. The implementation is in TensorFlow\texttrademark \cite{TF} and our Python code will be made publicly available.

\begin{table}[ht]
	\caption{Object recognition accuracies (percentage) for the 6 standard splits of the Office dataset. Each split represents a domain shift SOURCE $\rightarrow$ TARGET.}
	\label{tab:results}
	\centering
	\begin{tabular}{| r || c c c c c c|l|}
		\toprule
		& A $\rightarrow$ D & A $\rightarrow$ W & D $\rightarrow$ A & D $\rightarrow$ W & W $\rightarrow$ A & W $\rightarrow$ D  & Average \textit{(gain)}  \\
		\hline
		\hline
		AlexNet \cite{AlexNet}& 57.8  & 60.2 & 40.0 & 95.2 & 40.0 & 97.8 & 58.6\\
		Deep Coral \cite{DeepCORAL} & 58.9 & 65.9 & \textbf{40.7} & \textbf{95.6} & \textbf{41.6} & 98.0 & 60.5 \textit{(+1.9)}\\
		\textbf{\textit{Log D-Coral} } & \textbf{62.0} & \textbf{68.5}  & 40.6 & 95.3 & 40.6 & \textbf{98.7} & \textbf{61.4 \textit{(+2.8)}} \\
		\bottomrule
	\end{tabular}
\end{table}

\begin{table}[ht]
	\caption{Hyperparameters wighting the covariance losses. They had to be chosen differently for each domain shift, since each represents an independent problem, where domain adaptation is needed to a different extent. We found that, in general $\alpha$ has to be chosen approximately one order of magnitude higher than $\lambda$.}
	\label{tab:parameters}
	\centering
	\begin{tabular}{| l || c c c c c c|}
		\toprule
		Loss weight & A $\rightarrow$ D & A $\rightarrow$ W & D $\rightarrow$ A & D $\rightarrow$ W & W $\rightarrow$ A & W $\rightarrow$ D   \\
		\hline
		\hline
		$\lambda$ (\textit{Deep Coral \cite{DeepCORAL}}) & 1. & 1. & 0.05 & 0.05 & 0.1 & 0.1 \\
		$\alpha$ (\textit{Log D-Coral) } & 10. & 10. & 0.1 & 0.1 & 5. & 1.  \\
		\bottomrule
	\end{tabular}
\end{table}

\paragraph{Discussion}
Percentage classification accuracies are reported in Table \ref{tab:results}. The average percentage gain of Deep CORAL is consistent with the 2\% gain published in \cite{DeepCORAL}. The $\log$ loss introduced in this work contributes with an additional 1\% approximately. Our approach achieves better accuracies in 3 out of the 6 splits. However the margins are very small in the remaining cases, with respect to both the baseline and Deep CORAL.

The results of Table \ref{tab:results} prove i) that covariance alignment is indeed effective in filling the domain gap ii) that covariances must be regarded as matrices belonging to their natural embedding space, i.e.  $\mathrm{Sym}^{++}(n)$, and should thus be compared with appropriate distance measures.



\begin{figure}[t!]
	\centering
	\includegraphics[width=0.40\textwidth]{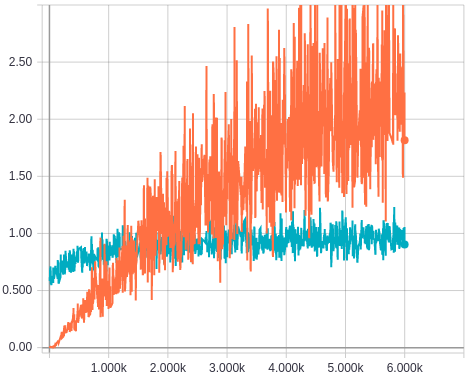} $\qquad$
	\includegraphics[width=0.40\textwidth]{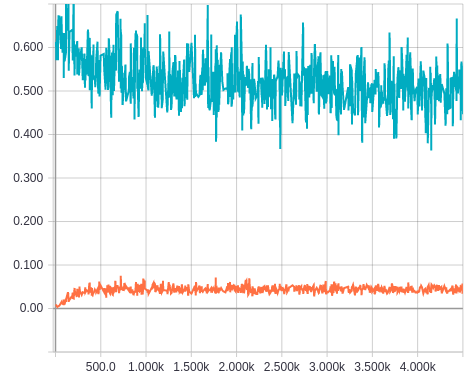} \\
	(a) $\qquad$ $\qquad$ $\qquad$ $\qquad$ $\qquad$ $\qquad$ $\qquad$ $\qquad$ (b)
	\caption{Batch-wise value of the terms $\alpha L_{log}$ (cyan) and $\lambda L_{CORAL}$ (orange) for the A $\rightarrow$ W split. (a) A standard training (\textit{no domain adaptation}), i.e. no distance constraint is enforced in the total loss function. (b) \textit{Domain adaptation}: $L_{log}$ and $L_{CORAL}$ are here added to the loss function and minimized jointly with the classification loss.}
	\label{fig:distances}
	
\end{figure}

In order to get a better understanding of the difference between Deep CORAL and Log-D-Coral we plot in Figure \ref{fig:distances} their weighted losses $\alpha L_{log}$ and $\lambda L_{CORAL}$ (as from equations (\ref{eq:D-CORAL}) and (\ref{eq:logLoss1}) for the domain shift A $\rightarrow$ W.
 
Figure \ref{fig:distances}(a) shows the batch-wise value of the two distance terms in a normal training, i.e. the two losses are calculated but no domain adaptation is enforced. Both losses naturally increase since the statistics of feature representations learned from the source are likely to diverge from the target ones, as the network specializes more and more to source data. However, $L_{CORAL}$ shows a pathological behavior, still increasing at an almost linear rate even when training reaches convergence and weights are only slightly updated. One would expect that little variations in the weights should instead produce little variations in the distance. On the contrary, $L_{log}$, even though increasing as expected, shows a way more reasonable trend. In fact, as training approaches convergence, $L_{log}$ tends to stabilize. Last, $L_{CORAL}$ is more noisy than $L_{log}$ (oscillation are bigger despite we plot $\lambda L_{CORAL}$, with $\lambda \ll \alpha$), meaning that its value can change a lot from one batch to another, which is a very undesirable property, meaning that the batch-wise distance is not well representative of the distance between the whole source and target datasets. This behavior can be interpreted as evidence of the sub-optimal nature of the Euclidean metric with respect to geodesic distances in $\mathrm{Sym}^{++}(n)$.

Figure \ref{fig:distances}(b) depicts instead the value of the two losses included in the minimization problem. As reported in \cite{DeepCORAL}, $L_{CORAL}$ experiences stabilization after increasing for a few epochs. This behavior is quite unclear, since we are trying to minimize it, but possibly depends on the base value of the distance on the uninitialized network. $L_{log}$, on the contrary, stabilizes within few epochs, after being minimized, which is somehow more reasonable. Oscillations are here comparable given that we plot $\alpha L_{log}$ and $\lambda L_{CORAL}$, with $\alpha = 10 \lambda$).

\section{Conclusion}
\label{sec:concl}
In this paper we have presented a natural extension of the end-to-end deep domain adaptation method proposed in \cite{DeepCORAL}. By rigorously accounting for the geometric properties of SPD matrices, we are able to improve object recognition performances for the Office dataset and prove the sub-optimality of the Euclidean metric in measuring distances between covariances.

As a possible future reasearch line, we would like to investigate Log-Hilbert-Schmidt distances between covariance operators. Although complex, such distances have been successfully exploited in an approximated form for image classification purposes \cite{Minh:CVPR16}.

{\small
	\bibliographystyle{ieee}
	\bibliography{egbib}
}

\end{document}